\documentclass[twocolumn]{article}

\usepackage{titlesec}
\titlespacing*{\section}
{0pt}{0.3cm}{0.3cm}

\titlespacing*{\subsection}
{0pt}{0.2cm}{0.2cm}

\newcommand{\secfnt}{\fontsize{11}{11}}
\newcommand{\ssecfnt}{\fontsize{9}{9}}

\titleformat{\section}
{\normalfont\secfnt\bfseries}{\thesection}{1em}{}
 
\titleformat{\subsection}
{\normalfont\ssecfnt\bfseries}{\thesubsection}{1em}{}

\usepackage[left=1.7cm, right=1.5cm, top=1.5cm, bottom=1.7cm]{geometry}
\usepackage{amsmath}
\usepackage{multirow}
\usepackage{latexsym}
\usepackage{subfigure} 
\usepackage{amssymb}
\usepackage{graphicx}
\usepackage{epstopdf}
\usepackage{color}
\usepackage{xcolor}
\usepackage[ruled,vlined]{algorithm2e}
\usepackage[utf8]{inputenc}
\usepackage{algorithmic}
\usepackage{lipsum}
\usepackage{fancyhdr}
\usepackage{authblk}
\usepackage{lipsum}  
\usepackage[labelfont=bf,textfont=it]{caption}

\definecolor{upmaroon}{rgb}{0.69, 0.11, 0.23}

\usepackage{graphicx}
\begin{document}
%
%*************************** HEADER *********************************
%
\title{
PageRank and The $K$-Means Clustering Algorithm
}
\author[1]{
Mustafa Hajij\thanks{e-mail: mustafahajij@gmail.com}
}
\author[2]{
Eyad Said\thanks{e-mail: esaid@mtmercy.edu}
}
\author[2]{
Robert Todd\thanks{e-mail: rtodd@mtmercy.edu}
}
\affil[1]{KLA Corporation}
\affil[2]{Mount Mercy University}
\setcounter{Maxaffil}{0}
\renewcommand\Affilfont{\itshape\small}
 \date{}

%\author{Mustafa Hajij\thanks{e-mail: mustafahajij@gmail.com}\\ %
%     	\scriptsize KLA Corporation %
%        \and 
%        Eyad Said\thanks{e-mail: esaid@mtmercy.edu}\\
%     	\scriptsize Mount Mercy University %     	
%        \and 
%        Robert Todd\thanks{e-mail: rtodd@mtmercy.edu}\\ %
%     	\scriptsize Mount Mercy University %
%}

%
%
\maketitle
%
%*************************** ABSTRACT *******************************
%

\textit{
\begin{abstract}
 We utilize the PageRank vector  to generalize the $k$-means clustering algorithm to directed and undirected graphs. We demonstrate that PageRank and other centrality measures can be used in our setting to robustly compute centrality of nodes in a given graph. Furthermore, we show how our method can be generalized to metric spaces and apply it to other domains such as point clouds and triangulated meshes. %Finally, we show that our approach can be utilized for point clouds clustering and mesh segmentation.
\end{abstract}
}
%
%******************************* BODY *********************************
%
%
%------------------
% SECTION: REVIEW
%------------------
%

\section{Introduction}
\label{sec:intro}
%Graphs are a ubiquitous data type encoding information about social networks, energy grids, computer networks, neural connections in our brains, etc. Yet, their abstractness makes graphs incredibly difficult to analyze.

%A good visual representation of a graph will present its \emph{structure} quickly and clearly, and enable further exploration. 

%Likely the most commonly used (though not necessarily the \emph{best}) node layout method, force-directed or mass-spring layouts~\cite{FruchtermanReingold1991}, rely on converting the graph into a physical system of springs and repulsive forces that iterative minimize an energy function. 
%These systems rely upon local relationships to reveal the overall shape in the graph. The result is an interactive method that finds structure in certain graphs, particularly sparse ones. However, this approach can, and often does, cause structures to overlap or cross paths, making them difficult to differentiate. 

The $k$-means algorithm is one of the most well-studied and popular point cloud clustering algorithms \cite{jain2010data}. The generic version of $k$-means algorithm takes as input a point cloud $X$ and the number of clusters $k$ and returns a partition of $X$ into $k$ subsets or clusters. Due to its popularity, the $k$-means algorithm has been studied extensively in the clustering literature and many variations of it have been suggested \cite{cheung2003k,chen2006k,dhillon2004kernel} including kernel versions \cite{dhillon2004kernel}. See also \cite{jain2010data} and the references therein.

This article addresses the graph clustering problem by leveraging centrality measures defined on the nodes of a given graph. More specifically, we utilize the \textit{PageRank} vector \cite{BrinPage1998} and view it as a centrality measure on a graph to generalize the $k$-means clustering algorithm to graphs. The algorithm introduced here is applicable to directed and undirectd graphs.

%For our purpose, utilizing the PageRank vector has multiple main advantages. First, PageRank is efficiently computed on very large graphs \cite{haveliwala1999efficient}. Second, as a scalar function on the nodes of the graph, the PageRank vector can be interpreted as a centrality measure for the nodes. Applications of the PageRank vector include citation analysis \cite{ma2008bringing}, image search \cite{jing2008pagerank}, and many others \cite{gleich2015pagerank}.

%More specifically, we view the PageRank as a continuous scalar function \cite{Pretto2008} defined on the vertices of the graph and utilize this scalar function to induce a \textit{filtration} as defined traditionally in the context of persistent homology. 
%We show that the \textit{persistence diagram} induced by this filtration can be utilized for graph similarity.

%The majority of graph visualization techniques rely on directly drawing nodes and edges, so called node-link diagrams, which require a layout method to place the nodes. The problem of automatic graph layout has a rich literature. 
%Many algorithms have been developed to achieve this task (i.e.~\cite{di1994algorithms}).
%Most of the proposed methods focus on finding an embedding of the graph with optimal readability,
%~\cite{tamassia1988automatic}, 
%relying on metrics such as symmetry of the graph, length of the edges, or number of edge crossings~\cite{erten2003simultaneous}.

%\subsection{Prior Work}

Graph clustering algorithms have a vast literature. The reader is referred to other sources for more details \cite{schaeffer2007graph}. Multiple graph clustering algorithms have been suggested over the past few decades including spectral-based methods \cite{dhillon2004kernel}, minimal spanning tree-based methods \cite{stuetzle2003estimating} and clique-based methods \cite{edachery1999graph}. See also \cite{schaeffer2007graph} for a other clustering methods on graphs. %A comparative study of graph clustering algorithms is given in \cite{brandes2003experiments}.

The $k$-means algorithm is related to the Voronoi diagrams which can be defined in the context of graphs \cite{zivanic2012voronoi}. Voronoi diagrams on graphs have been utilized for finding meaningful clusters in biological networks for example. \cite{zivanic2012voronoi}.

Several other works have found ways to apply the $k$-means algorithm (Lloyds's algorithm) to graphs of various flavors \cite{bota2015adaptations, witsenburg2011k}. As noted in \cite{erwig2000graph}, we computing the Voroni cells for directed and undirected graph is not particularly challenging despite the fact that we cannot define a metric for a general directed graph  (and thus applying the $k$-means algorithm). The main challenge in this context is the definition of the centroid of a cluster. The problem of defining the centroid was addressed with the Karcher/Fréchet mean \cite{grove1973conjugatec,pennec2006intrinsic} when considering point clouds in Riemannian manifolds. $K$-medoids clustering, as a discrete version of $k$-means \cite{rattigan2007graph}, is another method of addressing the problem of centroids in a discrete setting. These definitions are however very expensive to compute, in particular in the context of the $k$-means algorithm, and are not applicable directly to general directed graphs. The theory in \cite{chang2016mathematical} provides definitions that allow one to apply a $k$-means-like algorithm for computing clusters in any metric space without defining centroids. However, other definitions are required. 

%Here we take a completely phenomenological point of view, in that we apply a version of Lloyd's algorithm without reference to an objective function that we hope to minimize. We hope that our results warrant further investigations to the mathematical details of our proposed algorithm. 

Our novel contribution is to use nodes with high centrality measure in lieu of the centroid of a cluster in Lloyd's algorithm. While any centrality measure can be used, we focus on PageRank centrality. The algorithm we propose here is completely independent of any embedding of the graph into a metric space and is based only on the graph's connectivity structure. Nevertheless, we can apply this algorithm to graphs that are in a metric space such as a 3D-mesh or the neighborhood graph of point clouds in some Euclidean space.

Our algorithm has several main advantages. First, the PageRank vector can be defined for directed and undirected graphs. Second, that the PageRank vector was designed to be computed on massive graphs provides additional speed. Third, the algorithm we give here can be easily generalized to metric spaces making it applicable to other domains. Finally, the simplicity will be evident when we present the main algorithm.

\section{PageRank and Other Centrality Measures}

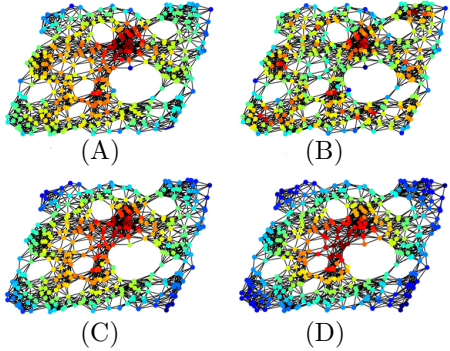
\begin{figure}[!h]
	\centering
	\includegraphics[width=0.7\linewidth]{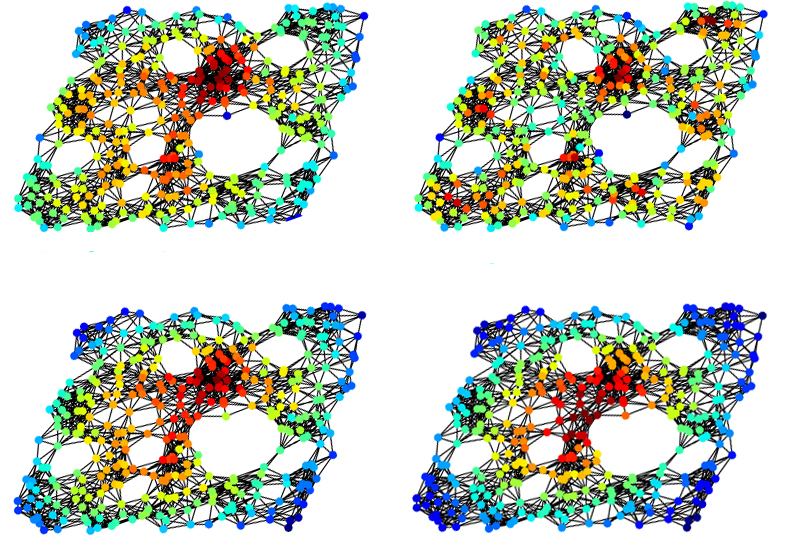}
	  \put(-145,59){(A)}
	  \put(-60,59){(B)}
	  \put(-145,-10){(C)}
	  \put(-60,-10){(D)}
	\caption{Various centrality measures on graphs. (A) Information centrality, (B) PageRank. (C) Harmonic centrality. (D) Closeness centrality.   }
	\label{fig.centrality}
\end{figure}

 The PageRank function \cite{BrinPage1998}  defined on the nodes of a graph can be viewed as centrality measure. 
For a directed graph $G(V,E)$, the PageRank function $PR:V \longrightarrow \mathbb{R}$ is defined for every vertex $v \in V$ by  $PR(v)=\frac{(1-\alpha)}{|V|}+\alpha \sum_{ u\in  out(v)} \frac{PR(u)}{ |out(u)|}$, 
where $out(v)$ is the set of nodes connected to $v$ by out edges leaving $v$; $0<\alpha<1$ is the \emph{damping factor}, typically set at $0.85$. When the graph is undirected, a different version of PageRank function is used  ~\cite{Grolmusz2012}.
The PageRank vector can be computed efficiently by the power method~\cite{HoffmanFrankel2018}. Intuitively, a high PageRank value at a given node $v$ usually means that $v$ is connected to many other nodes, which also have high PageRank scores. From this perspective, PageRank can be viewed as a measure of centrality for the nodes of the graph. See Figure \ref{fig.centrality} and observe that more central nodes in the graph example tend to have higher PageRank values (indicated with nodes with the red color).

While PageRank provides us with a good and fast measure of centrality for the nodes of the graph, other centrality measure can be utilized in Algorithm \ref{algo}. In fact, PageRank is a special case of a more general family of centrality measure called eigenvector centrality \cite{bonacich1987power} and these functions can also be used for this purpose.

Other centrality measures can also be utilized for our purpose. This includes harmonic centrality \cite{marchiori2000harmony,rochat2009closeness}, information centrality \cite{brandes2005centrality}, closeness centrality \cite{freeman1979centrality}, and VoteRank \cite{zhang2016identifying} among many other measures. Figure \ref{fig.centrality} shows a few example of centrality measures visualized on a graph.

\vspace{-10pt}
\section{The Main Algorithm}

Just like the traditional $k$-means algorithm, the algorithm we suggest here have three  stages : \textit{the initialization stage}, \textit{the assignment stage} and \textit{the update stage}. We discuss these  steps in details here. The termination of the algorithm can be chosen to be after a number of iterations or when the Voronoi diagrams do not update for a few consecutive iterations. A summary of the algorithm is given in Algorithm \ref{algo}. For the remainder of this section we assume that we are given a graph $G=(V,E)$, a method to compute the metric $d$ on $G$ and integer $k>0$ representing the number of desired clusters. We will not specify if the graph is directed or undirected as all methods introduced here are applicable to both types of graphs. 

\subsection{The initialization stage}

Just like the traditional $k$-means clustering algorithm, the algorithm we introduce here needs initial choice of the centroids. For our purpose here, we simply choose $k$ nodes from the graph $G$ uniformly at random. We will leave better initialization methods for future work.

%them uniformly, there are multiple method that one can utilize for choosing the initial $k$ nodes in the graph $G$. 

%Initialization of the $k$-means on point cloud using the using Voronoi diagram \cite{reddy2012initialization}. 

\subsection{The assignment stage}

The assignment stage starts with a list of $k$ centroids of the graph $G$. We denote these nodes by $c_1,\cdots c_k$. The metric $d$ along with the nodes $c_1,\cdots c_k$ induce a partition on $V$ called the \textit{Voronoi diagram of the graph} \cite{erwig2000graph}.

We recall quickly the definition of a Voronoi diagram on general metric spaces. Let $(X,d)$ be a metric space and let $C\subset X$ be subset of $X$, called the \textit{the subset of centroids}. The \textit{Voronoi cell} at point $c\in C$, denoted by $VC(c)$ is defined to be the set of all points $y\in X$  that are closer to $c$ than to any other point in $C$. The collection of subsets $VC(c)$ for all $c$ in $C$ is by definition the Voronoi diagram, denoted by $VD(C)$ of the metric space $X$ with respect to the subset $C$.

% Clearly, for any two points $c_1$ and $c_2$ in $C$, we have $V_{d}(c_1)\cap V_{d}(c_2)=\emptyset$. Moreover, $X=\cup_{c\in C }V_d(c).$

In the context of graphs, or other domains such as meshes and point clouds, any choice of a metric $d$ can be deployed and the Voronoi diagram can be computed using optimized algorithms depending on the domain of interest. Our version on graphs uses the algorithm given in \cite{erwig2000graph}. Finally, note that Voronoi diagrams can be computed for directed and undirected graphs. We refer the reader also to \cite{erwig2000graph} for details. In Algorithm \ref{algo}, at the end of the assignment stage the algorithm returns the Voronoi diagram $VD(\{c_1,\cdots,c_k\})$ of the centroids $c_1,\cdots,c_k$, which as we mentioned earlier consists of the sets $V_i:=VC(c_i) \subset V $ for each centroid $c_i$.

\subsection{The update stage}
The update stage assumes that we are given a partition of the node set : $V_1 \cdots V_k$. We use this partition to compute the subgraphs $G_i=(V_i,E_i)$ where $E_i=\{(u,v)\in E| u,v \in V_i\}$ for $1\leq i \leq k$. We then compute the PageRank $PR_i:V_i\longrightarrow\mathbb{R}$ for each $G_i$. The centroid of each graph $G_i$ is updated simply by $c_i :=argmax_{v\in V_i}(PR_i(v))$. In the rare case when the argmax function returns multiple centroids with the PageRank value, we choose one of these points arbitrarily. Notice the computation of the centroid with this method is reliant of the computation of the PageRank of the subgraphs. The PageRank vector can be computed very efficiently. See \cite{sarma2013fast} for a $\mathcal{O}( \sqrt{\log(n)}/\epsilon)$ distributed algorithm where $n$ is the number of nodes in the graph and $\epsilon$ is fixed constant.
%The main ingredient for the algorithm is the utilization of the centrality measures in the finding a central node.

\begin{algorithm}[h]
\label{algo}
\SetAlgoLined
\KwIn{Graph $G(V,E)$, number of clusters $k$.}
\KwOut{A partition of the node set $V$ into $k$ subsets.}
Initialize the set $C$ by choosing $k$ nodes from $V$\\
 \While{While termination criterion has not been met}{
  \For{$c_i$ in $C$
  }{Compute $V_i=VC(c_i)$}
   \For{ $V_i$ in $VD(C)$
  }{ Compute PageRank $PR_i$ on the subgraph $(V_i,E_i)$\\
  $c_i:=argmax_{v\in V_i}(PR_i(v))$
  }
 }
 \caption{PageRank-based $k$-means clustering algorithm on graphs.}
 \label{algo}
\end{algorithm}

\subsection{Distance computation}

In the computation of the Voronoi diagram one usually needs to compute the metric $d$ on $G$. It is important to notice that while the metric $d$ on $G$ is needed for this computation, one usually does not need to compute the entire distance matrix on $G$. 

In our experiments on graphs and triangulated meshes we utilized Dijkstra's algorithm \cite{Dijkstra1959} for the metric $d$. There are multiple methods to speed the distance computations on a graph \cite{potamias2009fast}. The heat method for computing geodesics introduced in \cite{crane2013geodesics} can also be utilized for fast metric computation on almost all domains that appear in practice.
Other metrics that depend on the graph Laplacian can also be utilized. This includes spectral type distances such as commute-time distance, discrete biharmonic distance, and diffusion distance.

\section{Extension of the Main Algorithm to Metric Spaces}
Algorithm \ref{algo} can be easily extended to metric spaces. Indeed, we notice that the initialization and assignment stages that compute the Voronoi cells can be defined for general metric spaces. It is in the update stage where PageRank is utilized. Given that PageRank is just a centrality measure, algorithm \ref{algo} generalizes to  metric spaces provided the computation of the PageRank function is replaced by an appropriate centrality measure. There are multiple centrality measures that satisfy this criterion e.g. the harmonic centrality and the closeness centrality.

% IS THIS NEEDED? There are multiple centrality measures that satisfy this criterion e.g. harmonic centrality

%\cite{marchiori2000harmony,rochat2009closeness}, and closeness centrality \cite{freeman1979centrality}. We leave further study of Algorithm \ref{algo} for future work.

%For instance, the \emph{commute-time distance} is defined as~\cite{FoussPirotteRenders2007} 
%\begin{equation}
%\label{commute}
%d_{ct}^2(x,y)=\sum_{i=1}^{|V|-1} \frac{1}{\lambda_i} (\phi_i(x)-\phi_i(y))^2. 
%\end{equation}
%Here $\{\lambda_i\}_{i=0}^{|V|-1}$ and $\{\phi\}_{i=0}^{|V|-1}$ are the generalized eigenvalues and eigenvectors of the graph Laplacian of $G_i$, respectively~\cite{chung1997spectral}. 
%In practice, we approximate the summations of Equation (\ref{commute})  by considering the first few nonzero eigenvectors, since the higher eigenvectors do not contribute significantly.

\section{Results}
To validate the results we applied the main algorithm on several datasets. 

\begin{figure}[h]
	\centering
	\includegraphics[width=0.6\linewidth]{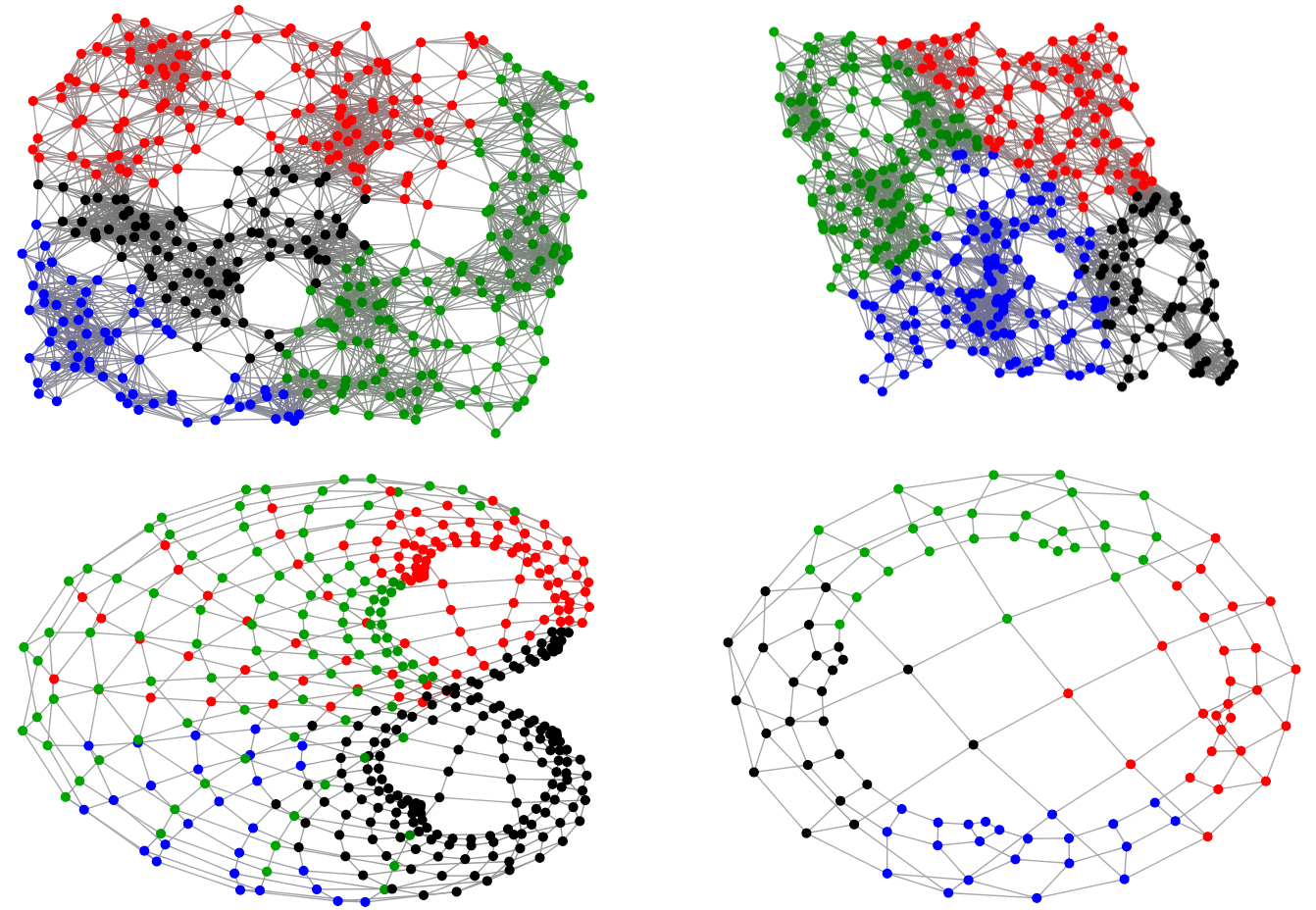}
	\caption{Applying Algorithm \ref{algo} on various graph examples. Number of clusters in all examples is $4$. The clusters are indicated by the colors of the nodes.}
	\label{fig.results}
\end{figure}

We applied our method to some of the graphs available in the NetworkX library \cite{hagberg2008exploring}. Figure \ref{fig.results} shows the application of Algorithm \ref{algo} on various graph examples. 

\vspace{-4pt}

\begin{figure}[h]
	\centering
	\includegraphics[width=0.8\linewidth]{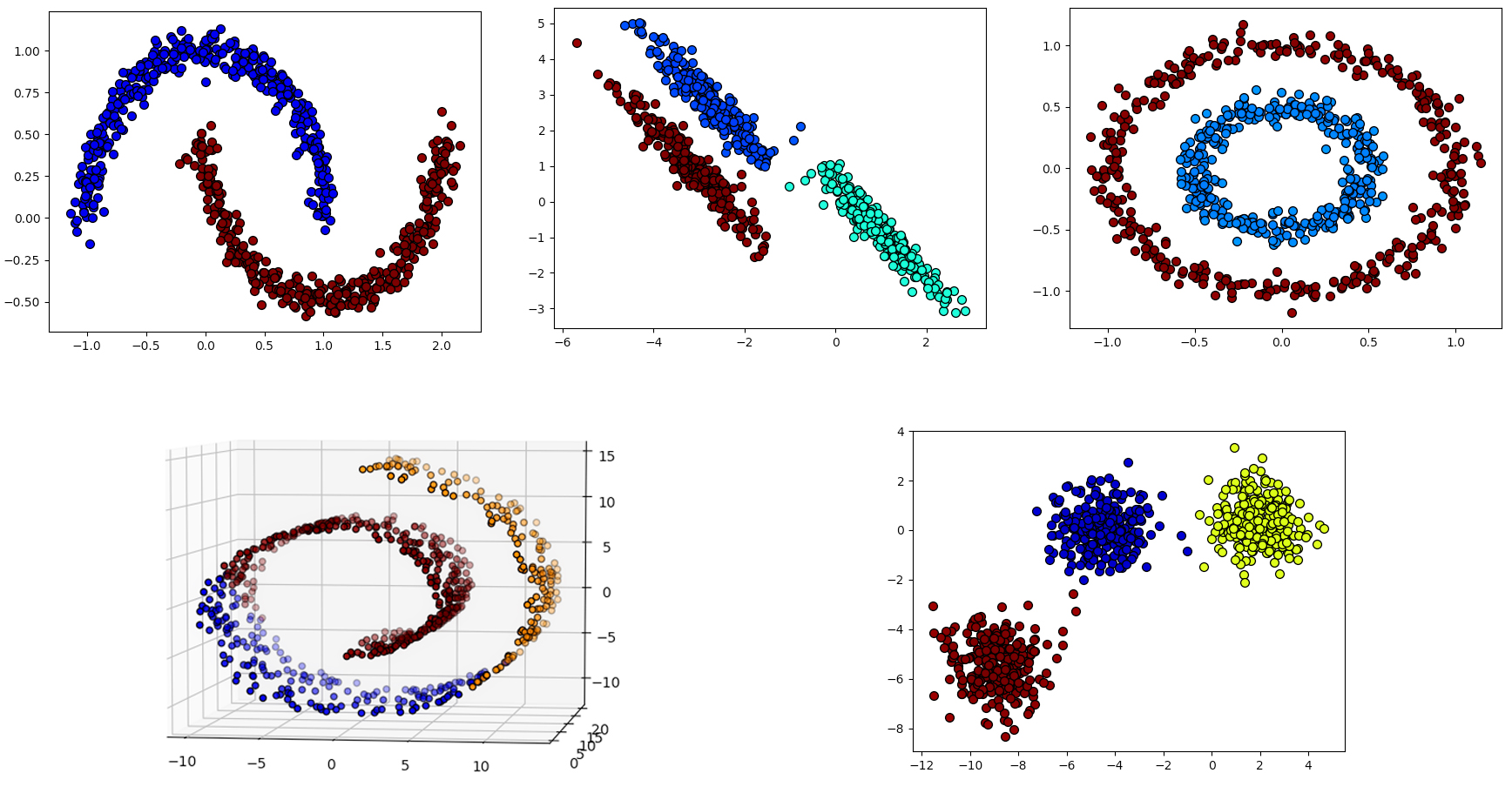}
	\caption{Various example of application of the algorithm on point cloud. For each point the following two steps are applied : (1) The neighborhood graph of the point cloud is calculated. (2) the graph $k$-means clustering algorithm is applied. The clusters are indicated by the colors of the points.}
	\label{fig.results_PC}
\end{figure}

The method that we proposed here is also applicable to point clouds from multiple perspectives. One way to do that is to compute a \textit{neighborhood graph} of the point cloud and then apply algorithm \ref{algo} on the graph to obtain a partition of the point cloud. We quickly recall the definition of a neighborhood graph. Let $S \subset \mathbb{R}^n$ be a point cloud with a distance function $d_S$ defined on $S$. Let $\epsilon>0$ be a positive number. The neighborhood graph is an undirected graph $\Gamma_{d_S,\epsilon}(S)$, where $\Gamma_{d_S,\epsilon}(S) = (S, E(\Gamma_{d_S,\epsilon}))$
and
$E(\Gamma_{d_S,\epsilon}) = \{[u, v] \mid d_S(u, v) \leq  \epsilon, u, v \in  S, u\neq v\}$. For our computation $d_S$ is simply the Euclidean metric. Figure \ref{fig.results_PC} shows the clusters obtained by applying Algorithm \ref{algo} on the neighborhood graph of  some point cloud examples. Note that the clusters obtained cannot be usually obtained using traditional $k$-means algorithm on point cloud with the usual Euclidean distance.

Finally, mesh segmenting can be considered as a clustering problem. For instance, in \cite{zhang2005mesh} a mesh segmentation is introduced via spectral clustering. See also \cite{shamir2008survey}.
 
\begin{figure}[!h]
	\centering
	\includegraphics[width=0.35\linewidth]{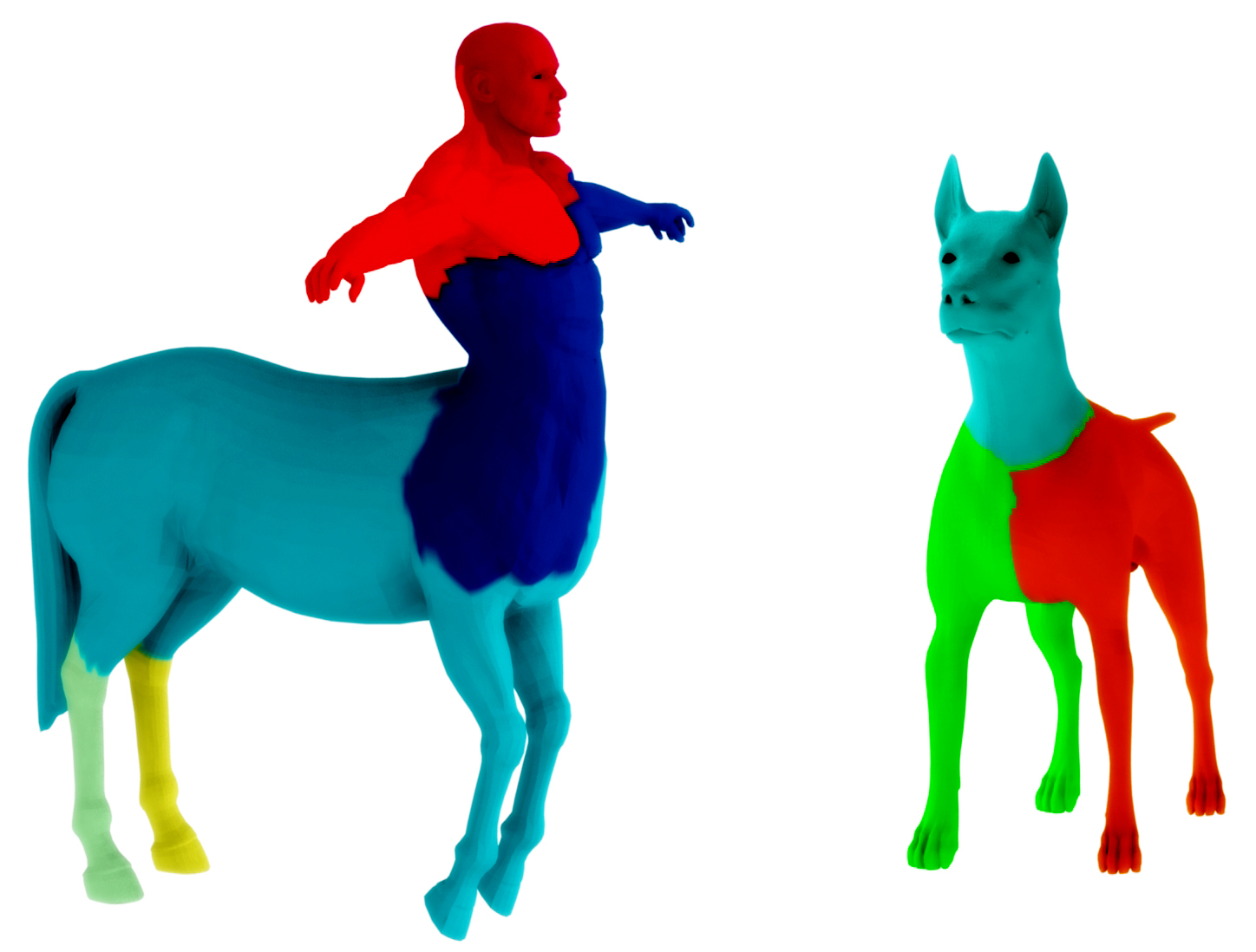}
	\caption{Viewed as metric spaces, we apply algorithm \ref{algo} on triangulated meshes. On the left the algorithm is applied with $6$ clusters and on the right the algorithm is applied with $3$ clusters.}
	\label{fig.results-pagerank}
\end{figure}

 In our context, we can view the mesh as a graph and apply Algorithm \ref{algo} immediately to this graph. A potentially better approach is to utilize a centrality measures that better describe the geometry of the underlying mesh such as harmonic centrality. Figure \ref{fig.results-pagerank} show examples of applying Algorithm \ref{algo} to triangulated meshes.

\vspace{-5pt}

\begin{figure}[h]
	\centering
	\includegraphics[width=0.6\linewidth]{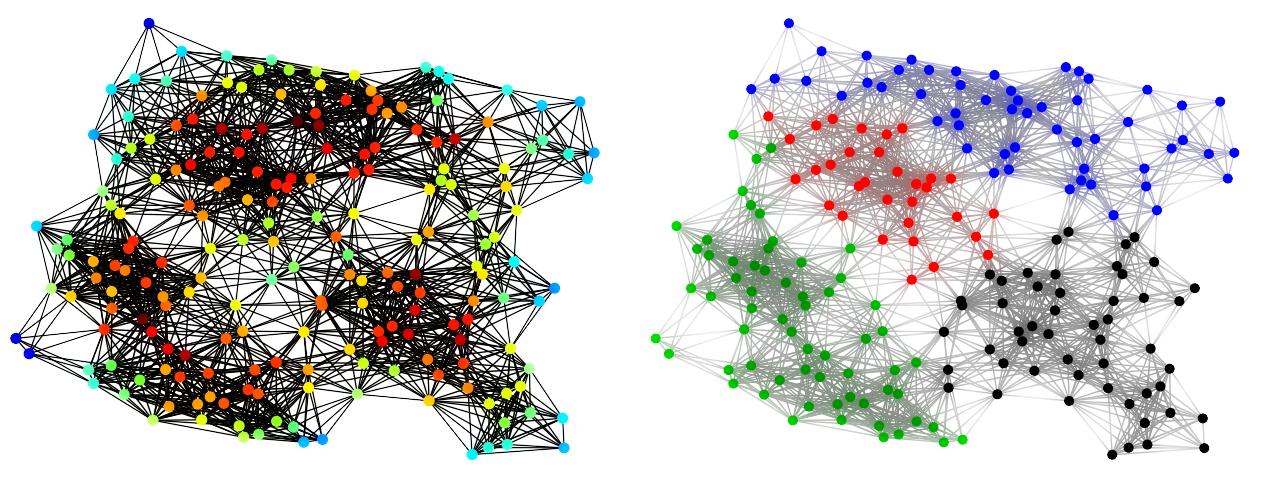}
	\caption{The PageRank function is utilized as a centrality measure in our work. The figure shows the visualization of the PageRank function on the nodes of the graph on the left. On the right we show the application of our algorithm on the same graph with $k=4$. The clusters are indicated by the colors of the nodes.}
	\label{fig.pagerank_and_graph}
\end{figure} 
\vspace{-5pt}

\section{Conclusion and Future Work}

%In this paper we introduced a graph clustering algorithm that generalizes $k$-means to graphs. 

The method introduced in this paper utilizes centrality measures such as PageRank to generalize the $k$-means clustering algorithm to graphs. While we explained quickly how our method is applicable to general metric spaces, we have not studied the theoretical properties of the algorithm in this context. Also, it is still not clear which centrality measure yields the best performance under a given metric. We are planning to pursue these directions in future work.
%\begin{figure}[h]
%	\centering
%	\includegraphics[width=0.4\textwidth,clip,trim={3.0in 2.5in 3.0in 2.5in}]{Figure.pdf}
%	\caption{A sample figure used in the body of the paper. Use coefficients less than half for textwidth to avoid images larger than the column size.}
%	\label{fig:fig1}
%\end{figure}

%\begin{figure*}[t!]
%	\centering
%	\includegraphics[width=0.6\textwidth,clip,trim={3.0in 2.5in 3.0in 2.5in}]{Figure.pdf}
%	\caption{A sample figure used in the body of the paper spanning both columns. Use coefficients less than the textwidth to avoid images larger than the page size.}
%	\label{fig:fig2}
%\end{figure*}

%You can also have a two column wide figure if you wish as illustrated in Figure \ref{fig:fig2}.

\bibliographystyle{plain}
\bibliography{network_tda_vis}
\end{document}